\documentclass[final,5p,times,twocolumn]{elsarticle}

\usepackage[colorlinks,
            linkcolor=blue,
            anchorcolor=blue,
            citecolor=blue]{hyperref}
\biboptions{sort&compress}
\usepackage{amssymb}
\usepackage{comment}
\usepackage{amsmath,amsfonts}
\usepackage{algorithmic}
\usepackage{algorithm}
\usepackage{array}
\usepackage[caption=false,font=normalsize,labelfont=sf,textfont=sf]{subfig}
\usepackage{textcomp}
\usepackage{stfloats}
\usepackage{url}
\usepackage{verbatim}
\usepackage{graphicx}
\usepackage{cite}
\usepackage{overpic} 
\usepackage{subcaption}
\usepackage{booktabs,multirow} 
\usepackage{amssymb} 
\usepackage{bbding} 
\usepackage{ulem} 

\begin{document}

\begin{frontmatter}



\title{Towards Context-aware Convolutional Network 
for Image Restoration}


\cortext[cor1]{Corresponding author.}

\author[address1,address2]{Fangwei Hao}
\ead{haofangwei@mail.nankai.edu.cn}

\author[address1,address2]{Ji Du}
\ead{1120230244@mail.nankai.edu.cn}

\author[address1,address2]{Weiyun Liang}
\ead{weiyunliang@mail.nankai.edu.cn}

\author[address1,address2]{Jing Xu \corref{cor1}}
\ead{xujing@nankai.edu.cn}

\author[address1,address2]{Xiaoxuan Xu \corref{cor1}}
\ead{xuxx@nankai.edu.cn}



\address[address1]{the College of Artificial Intelligence, Nankai University, Tianjin, 300350, China}
\address[address2]{Ocean Engineering Research Center, Nankai University, Tianjin 300350, China}

\begin{abstract}
Image restoration (IR) is a long-standing task to recover a high-quality image from its corrupted observation. Recently, transformer-based algorithms and some attention-based convolutional neural networks (CNNs) have presented promising results on several IR tasks. However, existing convolutional residual building modules for IR encounter limited ability to map inputs into high-dimensional and non-linear feature spaces, and their local receptive fields have difficulty in capturing long-range context information like Transformer. Besides, CNN-based attention modules for IR either face static abundant parameters or have limited receptive fields. To address the first issue, we propose an efficient residual star module (ERSM) that includes context-aware "star operation" (element-wise multiplication) to contextually map features into exceedingly high-dimensional and non-linear feature spaces, which greatly enhances representation learning. To further boost the extraction of contextual information, as for the second issue, we propose a large dynamic integration module (LDIM) which possesses an extremely large receptive field. Thus, LDIM can dynamically and efficiently integrate more contextual information that helps to further significantly improve the reconstruction performance. Integrating ERSM and LDIM into an U-shaped backbone, we propose a context-aware convolutional network (CCNet) with powerful learning ability for contextual high-dimensional mapping and abundant contextual information. Extensive experiments show that our CCNet with low model complexity achieves superior performance compared to other state-of-the-art IR methods on several IR tasks, including image dehazing, image motion deblurring, and image desnowing.
\end{abstract}




\end{frontmatter}

\section{Introduction}
The  goal of image restoration (IR) is to recover a high-quality image from a degraded one, in which there may exist noise, blur, or haze \citep{ref1}, \citep{ref2}. In order to address this long-standing and ill-posed problem, conventional methods have used numerous hand-crafted characteristics and assumptions to restrict the solution space. These methods, however, cannot be feasible in more complex real-world scenarios. In recent years, convolutional neural networks (CNNs) \citep{ref3} with strong learning abilities have greatly accelerated the advancement of image restoration and shown exceptional performance when compared with conventional methods \citep{ref4}, \citep{ref5}, \citep{ref6}. Then, numerous advanced units have been introduced or developed from other domains to improve IR performance, including the residual connection \citep{ref7}, the encoder-decoder design \citep{ref8}, dilated convolution \citep{ref9}, as well as attention mechanisms \citep{ref10}, \citep{ref11}. Recently, the state-of-the-art 
\begin{figure}[!t]
\centering
\includegraphics[width=3in]{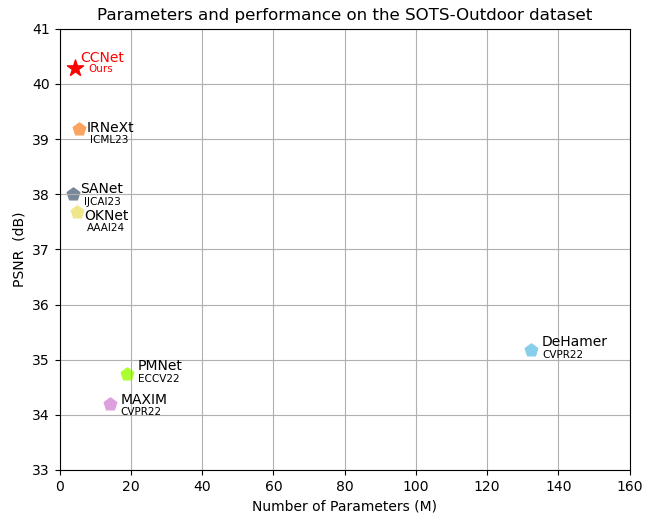}
\caption{Performance and parameters of different methods on SOTS-Outdoor dataset.}
\label{fig_1}
\end{figure}
performance of image restoration has been greatly advanced with the introduction of Transformer model \citep{ref12}. Nevertheless, using self-attention of Transformer to pursue large-scale receptive fields results in high model complexity due to the quadratic complexity of self-attention with respect to spatial resolution of images. In order to alleviate the complexity, researchers have proposed some certain algorithms to improve the IR efficiency of transformer-based methods. Concretely, both SwinIR \citep{ref13} and Uformer \citep{ref14} simplify the IR Transformer models by restricting the spatial zone of self-attention operation. Furthermore, to improve efficiency of capturing long-range pixel interactions, Restormer \citep{ref1} applies self-attention among the channel dimension rather than the spatial dimension. In addition, for image deblurring task, Stripformer \citep{ref15} presents a strip-based self-attention for simplification. Although these remedies do reduce the complexity of self-attention to some extent, they do not change its inherent properties; that is, their complexities remain quadratic with respect to the size of windows, channels, or strips.

 In contrast, the convolution operator with cheap computational cost has limited receptive fields and cannot capture long-range contextual information as Transformer can. To boost the IR performance of CNNs, numerous advanced networks \citep{ref16}, \citep{ref17}, \citep{ref18} have been proposed to strengthen the ability of CNN-based IR frameworks, in which the stacking of massive residual building modules for representation learning has gained significant popularity. Moreover, these advanced networks opt for the use of static extremely large convolution kernels to capture contextual information. Specifically, LKDNet \citep{ref16} seeks to obtain large receptive fields for image dehazing by decomposing a 21 × 21 static convolution into a smaller depth-wise convolution and a depth-wise dilated convolution. LaKDNet \citep{ref17} pursues large receptive fields by using large kernel convolutions (e.g., 9 × 9), followed by point-wise convolutions. Besides, MAN \citep{ref18} leverages a depth-wise convolution, a depth-wise dilated convolution, and a point-wise convolution to achieve a large receptive field of a static large kernel convolution. While these advanced methods achieve some performance gains, their static convolutional kernels struggle to cope with the common dynamic and non-uniform blurs in IR tasks. 
 
 In order to dynamically and effectively capture contextual information for image restoration, Cui et al. \citep{ref19} propose a strip attention module (SAM) to gather information from neighboring pixels in the same horizontal or vertical direction for each pixel. Such a SAM can integrate information from an expanded region, yet its receptive field is still constrained, leading to capturing insufficient context information.  Additionally, IRNeXt \citep{ref11} adopts a multi-scale module (MSM), an efficient local attention module (LAM) and massive stacked residual building modules to facilitate multi-scale learning and handle spatially-varying blurs. OKNet \citep{ref21} adopts an omni-kernel module and massive stacked residual building modules to learn global-to-local feature representations. Overall, existing CNN-based attention networks for image restoration have two defects that are not beneficial: (a) Massive stacked residual building modules with local effective receptive fields encounter limited ability to contextually map inputs into high-dimensional and non-linear feature spaces. (b) Their attention modules either have the static convolutional filters and cannot favorably handle the common dynamic and non-uniform blurs in IR tasks, or have limited direct receptive fields and cannot extract abundant contextual information.
 
To deal with these issues, we propose a context-aware convolutional network (CCNet) for efficient image restoration in this paper. Specifically, we first elaborately design an efficient residual star module (ERSM) as the building block which includes context-aware "star operation" (element-wise multiplication) to contextually map input features into exceedingly high-dimensional and non-linear feature spaces, resulting in efficient learning ability for contextual information. Furthermore, to boost the extraction of contextual information and handle the common dynamic and non-uniform blurs in IR tasks, we propose a large dynamic integration module (LDIM) which possesses an extremely large square receptive field to dynamically and efficiently integrate more contextual information that helps to significantly improve the reconstruction performance. The learnable weights involved in LDIM are dynamically learned from the input features by convolutional layers, and we further experientially employ multi-scale receptive fields across feature groups to handle degraded blurs with different sizes.

When equipped with the proposed ERSM and LDIM, our CCNet outperforms state-of-the-art algorithms on several image restoration tasks, including image dehazing, motion deblurring and desnowing. As Fig \ref{fig_1} shows, on the SOTS \citep{ref22} dataset for image dehazing, our CCNet with low complexity achieves the highest PSNR compared to recent advanced IR methods, including DeHamer \citep{ref23}, MAXIM \citep{ref24}, PMNet \citep{ref25}, SANet \citep{ref19}, IRNeXt \citep{ref11}, and OKNet \citep{ref21}.

The main contributions of this paper are as follows:
\begin{itemize}
\item{We propose an efficient residual star module (ERSM) which includes context-aware "star operation" (element-wise multiplication) to contextually map input features into exceedingly high-dimensional and non-linear feature spaces, and we take it as the basic building block of our network to extract representative contextual features.}

\item{We propose a large dynamic integration module (LDIM) which possesses an extremely large square receptive field. LDIM can dynamically and efficiently integrate more contextual information to significantly improve the reconstruction performance.}

\item{By integrating the proposed ERSM and LDIM into an U-shaped backbone, we propose a context-aware convolutional network (CCNet) for efficient image restoration.}
\item{Extensive experiments show that our CCNet outperforms previous state-of-the-art methods on several image restoration tasks.}
\end{itemize}

\section{RELATED WORK}
\subsection{Image Restoration}
Image restoration (IR) has garnered significant interest from both academic and industrial communities due to its crucial importance in numerous fields, such as photography, self-driving technology, and medical imaging. In terms of its inverse and ill-posed characteristics, various conventional methods have been proposed to limit the solution space, such as using hand-crafted features and a range of assumptions \citep{ref3}. Then, deep CNN-based algorithms \citep{ref26}, \citep{ref27}, \citep{ref28}, have been developed to significantly improve IR performance. In these networks, the U-shaped network is a common architecture for representation learning of hierarchical feature \citep{ref29}. In addition, many advanced modules, such as diverse attention modules \citep{ref10}, \citep{ref11}, universal skip connection \citep{ref7}, and effective dilated convolution \citep{ref9}, have been introduced or developed from high-level tasks. Furthermore, Transformer-based models \citep{ref12}, \citep{ref13}, \citep{ref14}, \citep{ref15}, \citep{ref23} have also been employed on low-level tasks to better capture long-range interdependence, and they have substantially improved the state-of-the-art performance of IR tasks. Specifically, Guo et al. \citep{ref23} first propose a DeHamer which introduces Transformer into image dehazing task. Next, Chen et al. \citep{ref30} develop a multi-scale projection transformer (MSP-Former) for image desnowing. In order to alleviate the quadratic complexity of Transformer, Cui et al. \citep{ref19} recently propose a strip attention module (SAM) to learn contextual information from an expanded region, which accelerates model performance on several IR tasks. They further propose an OKNet \citep{ref21} which adopts an omni-kernel module and massive stacked residual building modules to learn global-to-local feature representations. 

\begin{figure*}[!t]
\centering
\includegraphics[width=7in]{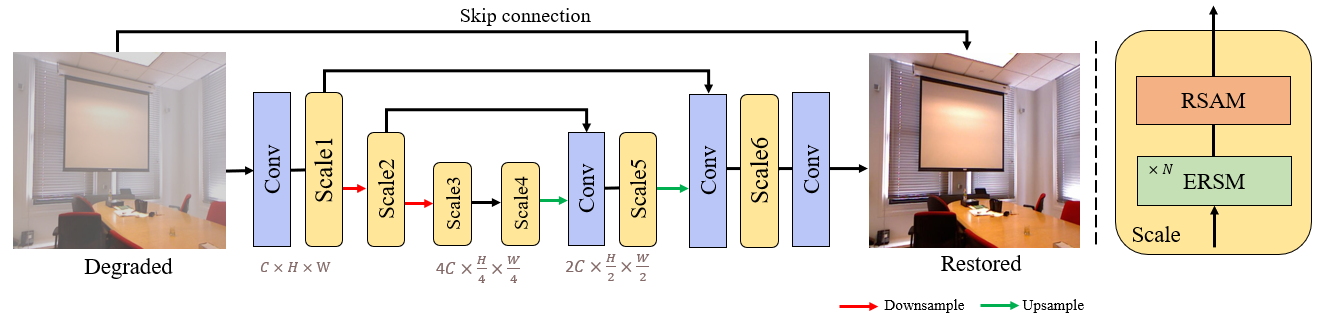}
\caption{Network architecture of our CCNet.}
\label{fig_2}
\end{figure*}

\subsection{Attention Mechanism}
In recent years, the attention mechanism has been widely applied in numerous tasks of computer vision. For image restoration, the aim of developed attention modules is to capture the interdependencies in spatial coordinates \citep{ref31}, along channels \citep{ref1}, \citep{ref32}, or both \citep{ref30}. For example, FFA-Net \citep{ref33} employs pixel attention and channel attention to flexibly handle various information. In order to efficiently fuse features, Liu et al. \citep{ref32} present a GridDehazeNet that adjusts the contributions of different feature maps based on channel-wise attention. Next, the feature filtering in MPRNet \citep{ref34} is implemented by the supervised attention module. These attention modules have significantly improved the performance of IR tasks.
Additionally, the development of efficient self-attention is a field that requires continuous research for image restoration (IR). Specifically, similar to Swin Transformer \citep{ref35}, both Uformer \citep{ref14} and SwinIR \citep{ref13} make use of the effective self-attention mechanism within local spatial regions. In contrast, Restormer \citep{ref1} focuses on channel self-attention instead of spatial dimension. To address motion blur, Stripformer \citep{ref15} has developed interlaced intra-strip and inter-strip attention layers to improve model performance. However, the complexities of these networks remain quadratic with respect to the size of windows, channels, or strips. In order to capture contextual information from an expanded region, a strip attention module (SAM) \citep{ref19} has been proposed and it demonstrates improved performance on several IR tasks. Nevertheless, SAM still faces limited receptive field which results in capturing insufficient context information. Besides, IRNeXt \citep{ref11} adopts a multi-scale module (MSM) and an efficient local attention module (LAM) to facilitate multi-scale learning and handle spatially-varying blurs.

\section{Proposed Method}
In this section, we firstly introduce the overall network architecture of proposed CCNet, then we give a detailed introduction to the proposed efficient residual star module (ERSM). Next, we present the proposed large dynamic integration module (LDIM) in detail. Later, we introduce the proposed residual star attention module (RSAM). Finally, we show the details of the loss function.

\subsection{Network Architecture}
As Fig \ref{fig_2} shows, a widely used encoder-decoder architecture is applied by our CCNet to learn hierarchical representations, and it is composed of six scales in total. Specifically, given a degraded input image with size $\mathbb{R}^{3\times H\times W}$, we utilize a single convolution layer to extract the shallow feature maps with size $\mathbb{R}^{H\times W\times C}$. Then, we feed the obtained shallow features into the encoder layers (Scale 1-3) for further refinement. During this process, the spatial resolution of the feature maps gradually decreases from $\mathbb{R}^{H\times W\times C}$ to $\mathbb{R}^{\frac H4\times\frac W4\times4C}$, while the number of channels increases stage by stage. At each stage, N efficient residual star modules (ERSMs) are stacked for contextual representation learning, followed by a proposed LDIM for dynamic contextual information aggregation and a strided convolution layer for downsampling operation. Next, to progressively recover the high-resolution feature maps, the obtained lowest-resolution feature maps are fed into the decoder layers (Scale 4-6) in which the spatial resolution of the feature maps gradually increases from $\mathbb{R}^{\frac H4\times\frac W4\times4C}$ to $\mathbb{R}^{H\times W\times C}$, while the number of channels decreases from 4C to C. In terms of feature upsampling, the transposed convolution is employed to genarate high-resolution feature maps. Following previous works \citep{ref1}, \citep{ref14}, \citep{ref19}, we adopt feature-level skip connections to deal with information loss due to downsampling operation. After concatenating the encoder features with corresponding decoder features, we use a standard convolution layer to fuse the feature maps. Finally, to produce the final sharp image, the original input image is added to yield the restored output, which forces the network to focus on learning the residual information. In addition, similar to previous methods \citep{ref7}, \citep{ref19}, \citep{ref36}, we also adopt multi-input and multi-output strategies to ease training.

\subsection{Efficient Residual Star Module (ERSM)}
\begin{figure*}[!t]
\centering
\includegraphics[width=6in]{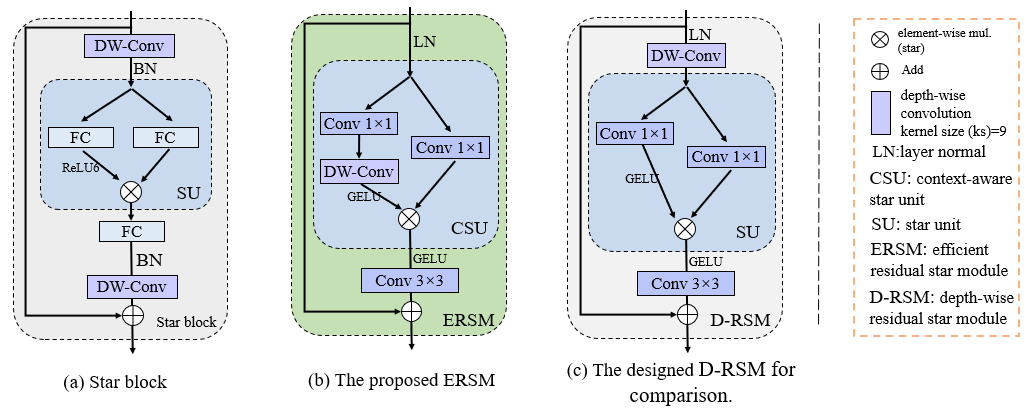}
\caption{The details of different modules. (a) Lightweight star block (StarB) in \citep{ref43} for high-level tasks. (b) The proposed efficient residual star module (ERSM) which includes context-aware star operation for low-level tasks. (c) The designed D-RSM for comparison.}
\label{fig_3}
\end{figure*}
In recent years, due to the widespread use of the self-attention mechanism, the element-wise multiplication has become a new paradigm to fuse features from different subspaces. For simplicity, we refer to this paradigm as ‘star operation’. Star operation has exhibited massive applications and powerful performance in the research fields of natural language processing (i.e., Monarch Mixer \citep{ref37}, Hyena Hierarchy \citep{ref38}, Mamba \citep{ref39}, etc.) and computer vision (i.e., HorNet \citep{ref40}, VAN \citep{ref41}, FocalNet \citep{ref42}, etc.). More recently, based on star operation, Ma et al. \citep{ref43} propose a simple yet powerful prototype, i.e., StarNet, in which they reveal that the star operation can map inputs into high-dimensional and non-linear feature spaces. As Fig \ref{fig_3}-(a) shows, they design a lightweight star block (StarB) as the building block of StarNet. While the StarB is capable of feature extraction for high-level tasks such as image classification, its star unit (SU) neglects to capture crucial contextual information, making it difficult to cope with complex input images of low-level tasks, incluing image dehazing, image deblurring, and image desnowing. In order for the residual module to map the input to high-dimensional spaces and learn contextual information, we propose an efficient residual star module (ERSM) for the image restoration. As shown in Fig \ref{fig_3}-(b), when the input features are sent into an ERSM module, the input features are first processed by the context-aware star unit (CSU) which contains two branches of convolutional operations. In a CSU, one of two branches includes a large depth-wise convolution to capture crucial contextual information. In contrast, as Fig \ref{fig_3}-(c) shows, we design a depth-wise residual star module (D-RSM) for comparison, which contains a same star unit (SU) as the StarB to further verify the efficiency of our CSU and ERSM. 

In an ERSM, given the input feature $\mathbf{X}_{input}$, it is processed firstly by the LN layer to obtain the normalized result $\mathbf{X}_{LN}$. We can express the process as
\begin{equation}
    \mathbf{X}_{LN}=LN(\mathbf{X}_{input})
\end{equation}
where $LN(\cdot)$ denotes the function of the LN layer. 
Then, $\mathbf{X}_{LN}$ is sent to two separate paths for processing, where one branch consists of a layer of 1×1 convolutional layers and the other branch consists of a 1×1 convolutional layer followed by a large depth-wise convolutional layer for capturing crucial contextual information, and a GELU activation function. Afterwards, the results of the two branches perform the star operation. The process can be formulated as
\begin{equation}
\small
\mathbf{X}_{star}=Conv_1(\mathbf{X}_{\boldsymbol{LN}})*GELU(Conv_{DW}(Conv_1(\mathbf{X}_{\boldsymbol{LN}})))
\end{equation}
where $Conv_1(\cdot)$,$Conv_{DW}(\cdot)$ and $GELU(\cdot)$ are the functions of 1×1 convolution, depth-wise convolution and GELU activation, respectively. Besides, * denotes element-wise multiplication, i.e., star operation, and $\mathbf{X}_{star}$ is the obtained feature after being processed by star operation. 
Finally, after $\mathbf{X}_{star}$ is activated by a GELU activation function to increase the nonlinear mapping capability, a standard 3×3 convolutional layer is used for feature refinement, and a skip connection is employed to produce the final output of ERSM. We can formulate this process as
\begin{equation}
\mathbf{X}_{\mathrm{ERSM}}=\mathbf{X}_{input}+Conv_3(GELU(\mathbf{X}_{star}))
\end{equation}
where $Conv_3(\cdot)$, $GELU(\cdot)$, and $\mathbf{X}_{\mathrm{ERSM}}$ are the standard $3\times3$ convolution operation, the GELU activation function, and the output of ERSM, respectively. It can be seen that the ERSM not only has a context-aware star operation to contextually map inputs to high-dimensional and non-linear feature spaces, but also possesses the strong learning capacity for image restoration.

\subsection{Large Dynamic Integration Module (LDIM)}

\begin{figure*}[!t]
\centering
\includegraphics[width=6.9in]{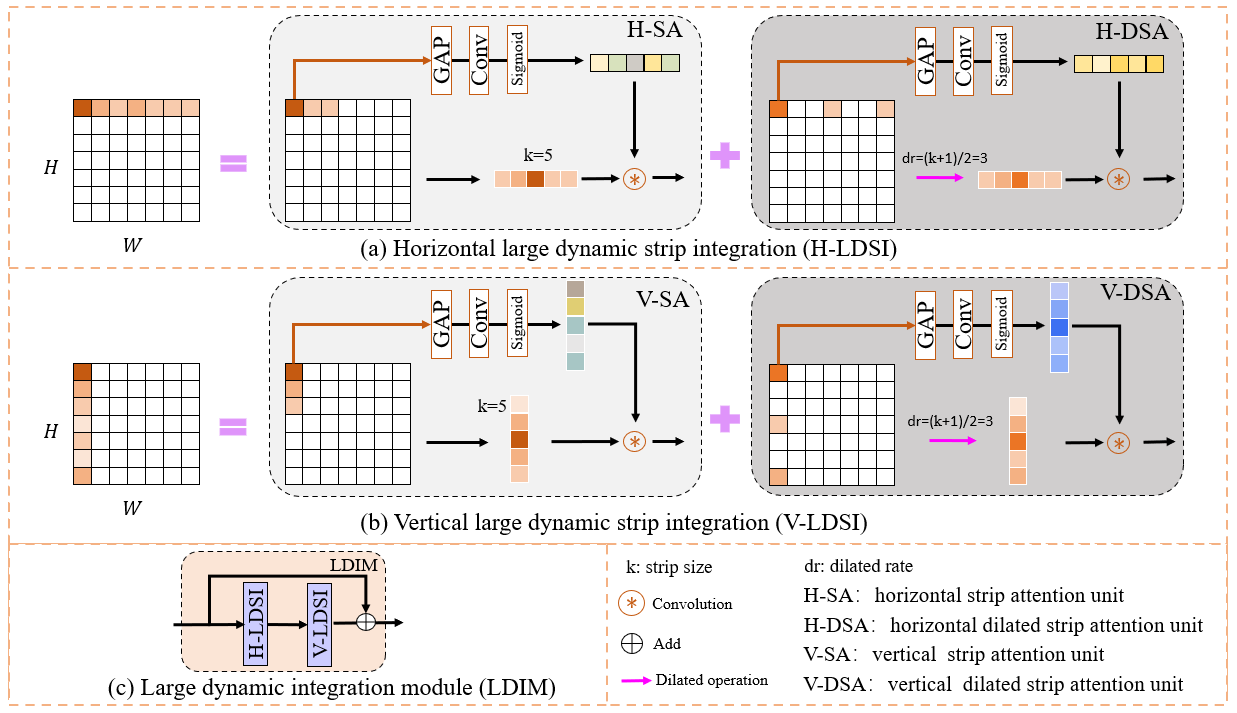}
\caption{The details of the proposed horizontal large dynamic strip integration (H-LDSI), vertical large dynamic strip integration (V-LDSI), and large dynamic integration module (LDIM).}
\label{fig_4}
\end{figure*}

Image restoration is a long-standing task that inevitably involves handling common dynamic and non-uniform blurs which standard convolution cannot favorably deal with. In order to cope with these complex inputs, Cui et al. \citep{ref19} propose a strip attention module (SAM) with learnable weights to capture contextual information from an expanded region, and it improves model performance on several IR tasks. However, the receptive field of SAM is still constrained, preventing it from effectively extracting contextual information from long-range pixels. To address this issue, as shown in Fig \ref{fig_4}-(c), we propose a large dynamic integration module (LDIM) which consists of a horizontal large dynamic strip integration (H-LDSI) module and a vertical large dynamic strip integration (V-LDSI) module. As Fig \ref{fig_4}-(a) shows, one H-LDSI module consists of a horizontal strip attention (H-SA) unit and a horizontal dilated strip attention (H-DSA) unit. Here, we have $\mathrm{dr}=(\mathrm{k+1})/2$, where dr represents the dilated rate of H-DSA and k denotes the strip size of H-SA. In one H-LDSI, we can obtain an extremely large strip receptive field for more contextual information. As the dr of the H-DSA increases, even though H-DSA discards numerous pixels within the dilated area, its operated pixels contain the contribution of any input pixel of the H-LDSI since these input pixels have already been integrated by H-SA. Thus, we use the H-LDSI to dynamically and effectively integrate more contextual information from an extremely large strip receptive field. Similarly, as Fig \ref{fig_4}-(b) shows, we employ the V-LDSI to integrate vertical information. Then, incorporating H-LDSI with V-LDSI, LDIM can dynamically and effectively integrate abundant contextual information from an extremely large square receptive field. 

Given the input feature $\mathbf{X}\in R^{H\times W\times C}$, where H×W and C  respectively denotes the spatial resolution and the number of channels, the process of one H-SA can be formulated as 
\begin{equation}
    \mathbf{X}_{h,w,c}^{\prime}=\sum_{k=0}^{K-1}A_{k}^{\prime}\mathbf{X}_{h,w-\lfloor\frac{K}{2}\rfloor+k,c}
\end{equation}
where $\mathbf{A}^{\prime}\in{R}^K$ is the attention weights which are dynamically learned from the input features by convolutional layers, and K is the strip size. To simplify the expression, we can formally express the process as
\begin{equation}
    \mathbf{X}_H^{\prime}=F_{H-SA}^K(\mathbf{X})
\end{equation}                                                                                
where $\mathbf{X}_H^{\prime}$ is the output of the H-SA, $F_{H-SA}^K(\cdot)$ denotes the process of H-SA. In addition, the process of H-DSA can be formulated as 
\begin{equation}
    \mathbf{X}_{h,w,c}^{\prime\prime}=\sum_{k=0}^{K-1}A_{k}^{\prime\prime}\mathbf{X}_{h,\left(w+k-\left[\frac{K}{2}\right]\right)*d,c}^{\prime},d=1,\ldots,D,
\end{equation}                                                      
where $\mathbf{A}^{\prime\prime}\in R^K$ is the obtained dynamic attention weights and d denotes the dilated rate (dr). Similarly, for expression simplification, we can formally express the process of H-DSA as 
\begin{equation}
    \mathbf{X}_H^{\prime\prime}=F_{H-DSA}^K(\mathbf{X}_H^\prime)
\end{equation}                                                                     
where $\mathbf{X}_H^{\prime\prime}$ denotes the output feature of the H-DSA. Thus, the process of one H-LDSI can be formulated as
\begin{equation}
    \mathbf{X}_H^{\prime\prime}=F_{H-DSA}^K(F_{H-SA}^K(\mathbf{X}))
\end{equation}                                                                              
where $F_{H-SA}^K(\cdot)$ and $F_{H-DSA}^K(\cdot)$ are the functions of H-SA and H-DSA, respectively. Similarly, the process of V-LDSI can be formulated as
\begin{equation}
    \mathbf{X}_V^{\prime\prime}=F_{V-DSA}^K(F_{V-SA}^K(\mathbf{X}_H^{\prime\prime}))
\end{equation}                                                                              
where $\mathbf{X}_H^{\prime\prime}$ and $\mathbf{x}_V^{\prime\prime}$ are the input and output of the V-LDSI, respectively, and $F_{V-SA}^{K}(\cdot)$ and $F_{V-DSA}^K(\cdot)$ are the processes of V-SA and V-DSA, respectively. Finally, the process of large dynamic integration module (LDIM) can be formulated as

\begin{equation}
    \small
    \mathbf{X}_{LDIM}=\mathbf{X}+F_{V-DSA}^K(F_{V-SA}^K(F_{H-DSA}^K(F_{H-SA}^K(\mathbf{X}))))
\end{equation}                                                    
where $\mathbf{X}_{LDIM}$ is the output features of the entire LDIM.

\subsection{Residual Star Attention Module (RSAM)} 
\begin{figure}[!t]
\centering
\includegraphics[width=3.5in]{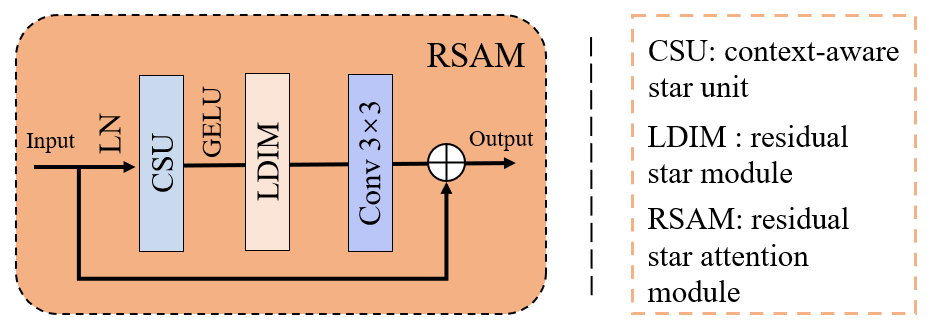}
\caption{The details of the proposed context-aware residual star attention module (RSAM).}
\label{fig_5}
\end{figure}

In order to efficiently and dynamically capture representative features, we further propose a residual star attention module (RSAM) by incorporating the context-aware star unit (CSU), large dynamic integration module (LDIM), a standard 3×3  convolution layer with residual structure. Thus, as Fig \ref{fig_5} shows, RSAM can not only contextually map features into an high-dimensional and non-linear feature space to extract discriminative features, but also dynamically and effectively integrate abundant contextual information from an extremely large square receptive field, making it possess strong learning capacity to cope with the complex inputs of image restoration. These advantages of RSAM significantly boost model performance.

\subsection{Loss Functions}
Following \citep{ref7}, to ease network training, the dual-domain $\mathrm{L}_1$ loss is employed to refine feature in both spatial and frequency domains. Thus, for each output, we can formulate the loss function as
\begin{equation}
    \begin{gathered}
L_{s}=\frac{1}{s}\parallel\hat{\mathbf{I}}-\mathbf{I}\parallel, \\
L_{f}=\frac{1}{s}\parallel\mathcal{F}\big(\hat{\mathbf{I}}\big)-\mathcal{F}(\mathbf{I})\parallel, \\
\mathrm{L}=L_{s}+\lambda L_{f} 
\end{gathered}
\end{equation}
where $\hat{\mathbf{I}}$ denotes the output images of the entire network and $\text{I}$ is the ground truth; $\mathcal{F}(\cdot)$ is the fast Fourier transform (FFT), and S denotes the total elements for normalization. During training process, we set $\lambda$ to 0.1 and the Adam optimization algorithm \citep{ref44} is adopted to optimize the entire network.

\section{EXPERIMENTAL RESULTS}
In order to verify effectiveness of our CCNet, we conduct extensive experiments on several image restoration (IR) tasks: image dehazing (RESIDE \citep{ref22}), image motion deblurring (GoPro \citep{ref45} and HIDE \citep{ref46}), and image desnowing (CSD \citep{ref47}). In this section, we first present the training settings. Then, based on these datasets, we provide the tested results and the corresponding illustration. Next, we conduct a series of ablation experiments to verify effectiveness of the proposed components and models. Lastly, we carry out the model complexity analysis.

\subsection{Settings}
During training, the Adam \citep{ref44} optimizer with $\beta_1=0.9$ and $\beta_2=0.999$ is adopted to optimize separate models for different IR datasets. The initial learning rate is set to $1e^{-4}$, and it decays to $1e^{-6}$ with the cosine annealing strategy. We set the batch size to 8 for the RESIDE-Outdoor \citep{ref22} dataset, and the batch size is set to 4 for other datasets. The data patches with size of 256×256 are applied to training models and we only adopt horizontal flips for data augmentation. In each scale of models, we employ different numbers of residual blocks N for different IR tasks. Specifically, according to the complexity of different tasks, we set N to 3 for image dehazing, 15 for image motion deblurring, and 5 for image desnowing. Besides, we conduct all experiments using the Pytorch \citep{ref48} framework with an NVIDIA 3090 GPU.

\subsection{Main Results}
\begin{figure*}[!t]
\centering
\includegraphics[width=6.5in]{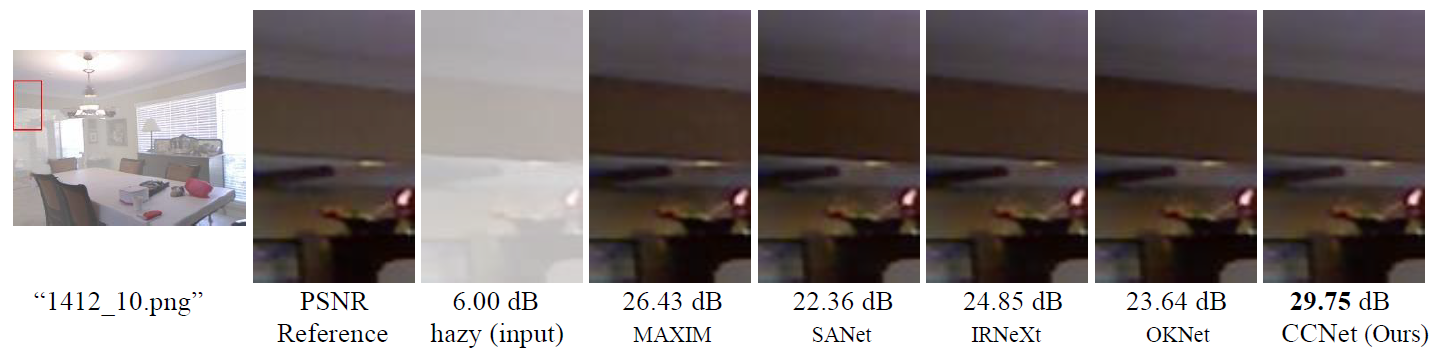}
\caption{Comparisons of Image dehazing on the SOTS-Indoor. The best result is highlighted.}
\label{fig_6}
\end{figure*}

\begin{figure*}[!t]
\centering
\includegraphics[width=6.5in]{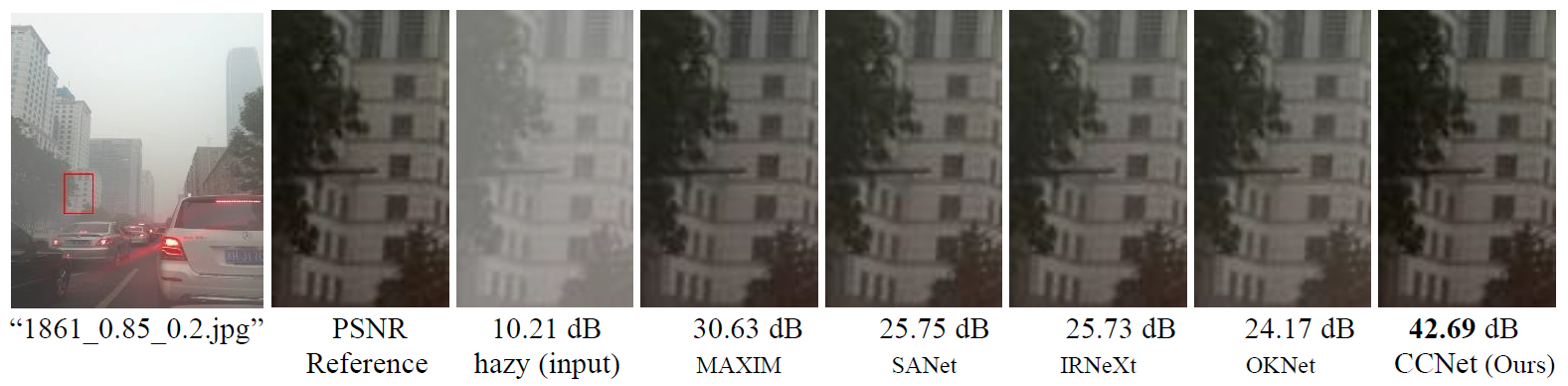}
\caption{Comparisons of Image dehazing on the SOTS-Outdoor. The best result is highlighted.}
\label{fig_7}
\end{figure*}

\textbf{Image dehazing.} The RESIDE \citep{ref22} dataset is adopted for training our CCNet, while the SOTS \citep{ref22} dataset is utilized for testing. The detailed testing results are presented in Table \ref{tab1}. Furthermore, we compare our CCNet with other state-of-the-art IR methods, including DCP \citep{ref49}, GCANet \citep{ref50}, GridDehazeNet \citep{ref32}, MSBDN \citep{ref51}, PFDN \citep{ref52}, FFA-Net \citep{ref33}, AECR-Net \citep{ref53}, MAXIM \citep{ref24}, DeHamer \citep{ref23}, PMNet \citep{ref25}, DehazeFormer-L \citep{ref54}, SANet \citep{ref19}, IRNeXt \citep{ref11}, and OKNet \citep{ref21}. As illustrated in Table \ref{tab1}, our CCNet yields the highest quantitative result 41.25 dB PSNR on the SOTS-Indoor dataset and it outperforms the recent transformer-based algorithm DehazeFormer-L by 1.2 dB PSNR while having only 17\% parameters and 16\% MACs of the latter. As for the comparison on SOTS-Outdoor dataset, the highest performance 40.29 dB PSNR is achieved by our CCNet with low complexity, demonstrating the superiority of our method. Specifically, compared to the expensive Transformer model DeHamer, our CCNet, with just 3.2\% parameters and 89\% MACs of it, achieves a 5.11 dB PSNR gain. In terms of the visual comparison, we carry out testing experiments on “1412\_10.png” of SOTS-Indoor dataset, and we illustrate the visual results and corresponding quantitative results of different methods in Fig \ref{fig_6} . Visual results demonstrate that our CCNet produces sharper details and yields better outcomes for image dehazing. Besides, we evaluate and record the corresponding quantitative PSNR results which show the consistency with visual comparison. In addition, in Fig \ref{fig_7} , we present the comparisons of different methods on the “1861\_0.85\_0.2.jpg” of SOTS-Outdoor dataset, and the visual and quantitative results reveal the similar conclusion to the comparison results of SOTS-Indoor dataset. Based on these experimental results, we can find that our CCNet is more effective and efficient when applied to the image dehazing task, producing clearer images which are visually closer to the reference images than other state-of-the-art algorithms.

\begin{table*}[htbp]
  \centering
  \caption{Image dehazing results on SOTS [Li et al., 2018]. The best and the second best results
 are \textbf{hightlighted} and \underline{underlined}, respectively.
}
    \begin{tabular}{lllllll}
    \toprule
    \multicolumn{1}{l}{\multirow{2}[4]{*}{Method}} & \multicolumn{2}{p{9.09em}}{\centering SOTS-Indoor} & \multicolumn{2}{p{9.09em}}{\centering SOTS-Outdoor} & \multicolumn{2}{p{9.09em}}{\centering Overhead} \\
\cmidrule{2-7}    \multicolumn{1}{r}{} & \multicolumn{1}{p{4.045em}}{PSNR↑} & \multicolumn{1}{p{4.545em}}{SSIM↑} & \multicolumn{1}{p{4.545em}}{PSNR↑} & \multicolumn{1}{p{4.545em}}{SSIM↑} & \multicolumn{1}{p{4.745em}}{Params (M)} & \multicolumn{1}{p{4.545em}}{MACs (G)} \\
    \midrule
    DCP   & 16.62 & 0.818 & 19.13 & 0.815 & \multicolumn{1}{p{4.045em}}{\ \ \ -} & \multicolumn{1}{p{4.045em}}{\ \ \ -} \\
    GCANet & 30.23 & 0.98  & \multicolumn{1}{p{4.045em}}{\ \ \ -} & \multicolumn{1}{p{4.045em}}{\ \ \ -} & 0.702 & 18.41 \\
    GridDehazeNet & 32.16 & 0.984 & 30.86 & 0.982 & 0.956 & 21.49 \\
    MSBDN & 33.67 & 0.985 & 33.48 & 0.982 & 31.35 & 41.54 \\
    PFDN  & 32.68 & 0.976 & \multicolumn{1}{p{4.045em}}{\ \ \ -} & \multicolumn{1}{p{4.045em}}{\ \ \ -} & 11.27 & 50.46 \\
    FFA-Net & 36.39 & 0.989 & 33.57 & 0.984 & 4.456 & 287.8 \\
    AECR-Net & 37.17 & 0.99  & \multicolumn{1}{p{4.045em}}{\ \ \ -} & \multicolumn{1}{p{4.045em}}{\ \ \ -} & 2.611 & 52.2 \\
    MAXIM & 38.11 & \underline{0.991} & 34.19 & 0.985 & 14.1  & 108 \\
    DeHamer & 36.63 & 0.988 & 35.18 & 0.986 & 132.45 & 48.93 \\
    PMNet & 38.41 & 0.99  & 34.74 & 0.985 & 18.9  & 81.13 \\
    DehazeFormer-L & 40.05 & \textbf{0.996} & \multicolumn{1}{p{4.045em}}{\ \ \ -} & \multicolumn{1}{p{4.045em}}{\ \ \ -} & 25.44 & 279.7 \\
    SANet & 40.40  & \textbf{0.996} & 38.01 & 0.995 & 3.81  & 37.26 \\
    IRNeXt & \underline{41.21} & \textbf{0.996} & \underline{39.18} & \underline{0.996} & 5.46  & 41.95 \\
    OKNet & 40.79 & \textbf{0.996} & 37.68 & 0.995 & 4.72  & 39.67 \\
    CCNet (ours) & \textbf{41.25} & \textbf{0.996} & \textbf{40.29} & \textbf{0.997} & 4.26  & 43.51 \\
    \bottomrule
    \end{tabular}%
  \label{tab1}%
\end{table*}%

\begin{figure*}[!t]
\centering
\includegraphics[width=6.5in]{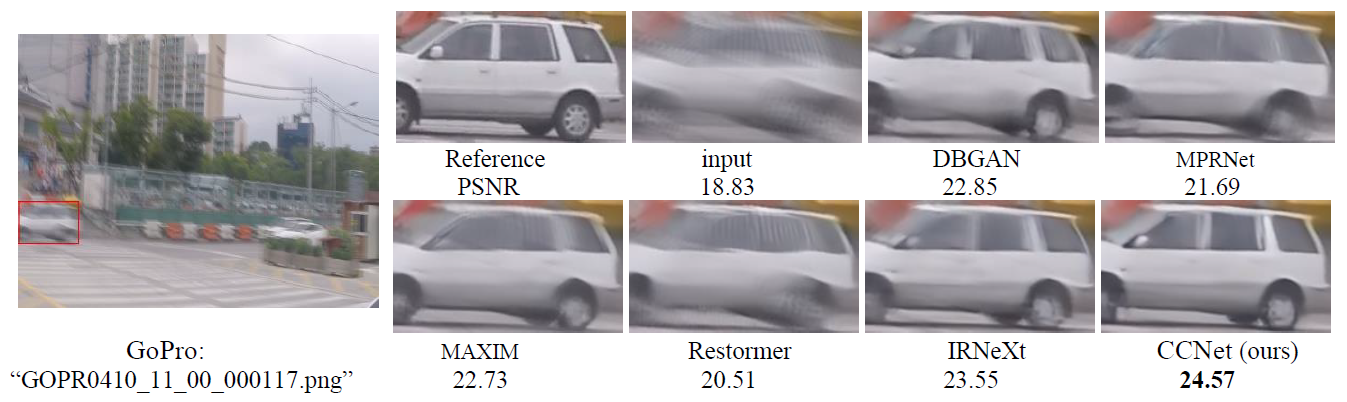}
\caption{Comparisons of image motion deblurring on the GoPro dataset. The best result is highlighted.}
\label{fig_8}
\end{figure*}

\textbf{Image motion deblurring.} We further evaluate our CCNet using the synthetic datasets to provide a comprehensive validation for single-image motion deblurring task. Concretely, the GoPro dataset \citep{ref45} is employed to train and test the obtained model, and the HIDE dataset \citep{ref46} is also chosen for testing. During the testing process, we use PSNR and SSIM as the evaluation metrics of model performance. Then, we further compare the results of our method with those of other state-of-the-art methods, including DeblurGAN-v2 \citep{ref55}, DBGAN \citep{ref56}, DMPHN \citep{ref57}, SPAIR \citep{ref58}, MIMO-UNet+ \citep{ref59}, IPT \citep{ref12}, MPRNet \citep{ref31}, HINet \citep{ref60}, and Restormer \citep{ref1}. The detailed comparison results are presented in Table \ref{tab2}. On GoPro testing dataset, it can be seen that our CCNet outperforms the powerful transformer model Restormer by 0.39 dB PSNR and achieves 0.963 SSIM which is higher than the 0.961 SSIM of Restormer. As for the HIDE dataset, our CCNet obtains the second best PSNR 31.09 dB which is a little lower than the 31.22 dB of the Restormer, yet our obtained 0.945 SSIM is the highest SSIM which is higher than the second highest 0.942 SSIM of the Restormer, indicating that our CCNet can better recover the structural details in the image. Next, we further illustrate the visual results in Fig \ref{fig_8}. Compared with other competing methods, the proposed CCNet can not only yield clearer contours but also generate more detailed information of the foreground and background. These visual results indicate that our CCNet can recover more faithful outputs due to more contextual information extracted from the large receptive field.

\begin{table*}[htbp]
  \centering
  \caption{Image motion deblurring results on the GoPro and HIDE datasets, the best
and the second best results are \textbf{hightlighted} and \underline{underlined}, respectively.
}
    \begin{tabular}{p{6.345em}cccc}
    \toprule
    \multirow{2}[4]{*}{Method} & \multicolumn{2}{p{8.09em}}{\centering GoPro} & \multicolumn{2}{p{8.09em}}{\centering HIDE} \\
\cmidrule{2-5}    \multicolumn{1}{r}{} & \multicolumn{1}{p{4.045em}}{\centering PSNR↑} & \multicolumn{1}{p{4.045em}}{\centering SSIM↑} & \multicolumn{1}{p{4.045em}}{\centering PSNR↓} & \multicolumn{1}{p{4.045em}}{\centering SSIM↓} \\
    \midrule
    DeblurGAN-v2 & 29.55 & 0.934 & 26.61 & 0.875 \\
    DBGAN & 31.1  & 0.942 & 28.94 & 0.915 \\
    DMPHN & 31.2  & 0.94  & 29.09 & 0.924 \\
    SPAIR & 32.06 & 0.953 & 30.29 & 0.931 \\
    MIMO-UNet+ & 32.45 & 0.957 & 29.99 & 0.93 \\
    IPT   & 32.52 & \multicolumn{1}{p{4.045em}}{\ \ \ \ \ -} & \multicolumn{1}{p{4.045em}}{\ \ \ \ \ -} & \multicolumn{1}{p{4.045em}}{\ \ \ \ \ -} \\
    MPRNet & 32.66 & 0.959 & 30.96 & 0.939 \\
    HINet & 32.71 & 0.959 & 30.32 & 0.932 \\
    Restormer & 32.92 & 0.961 & \textbf{31.22} & \underline{0.942} \\
    IRNeXt & \underline{33.16} & \underline{0.962} & 30.91 & 0.94 \\
    \midrule
    CCNet (ours) & \textbf{33.31} & \textbf{0.963} & \underline{31.09} & \textbf{0.945} \\
    \bottomrule
    \end{tabular}%
  \label{tab2}%
\end{table*}%

\begin{figure*}[!t]
\centering
\includegraphics[width=6.5in]{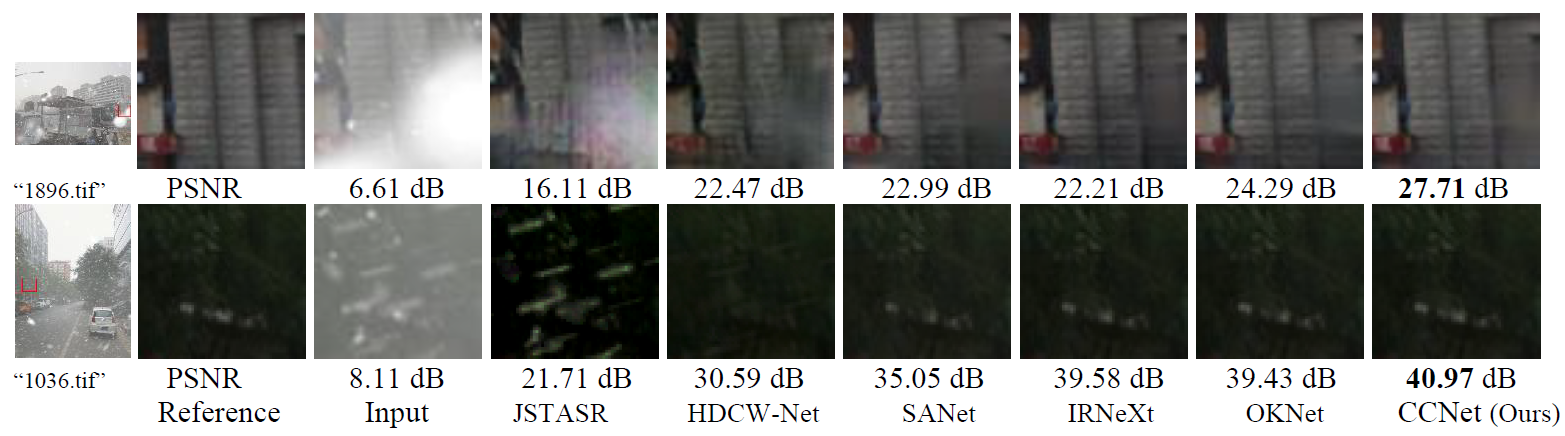}
\caption{Comparisons of Image desnowing on the CSD dataset. The best results are \textbf{highlighted}.}
\label{fig_9}
\end{figure*}

\textbf{Image desnowing.}  
To verify the effectiveness of our CCNet on image desnowing, we train and test our model on the CSD \citep{ref48} dataset. We employ PSNR and SSIM as the evaluation metrics of model performance, and we further compare our testing results with other state-of-the-art methods, including DesnowNet \citep{ref61}, CycleGAN \citep{ref62}, All in One \citep{ref63}, JSTASR \citep{ref64}, HDCW-Net \citep{ref47}, TransWeather \citep{ref65}, MSP-Former \citep{ref30}, NAFNet \citep{ref66}, SANet \citep{ref19} , IRNeXt \citep{ref11}, and OKNet \citep{ref21}. Table \ref{tab3} presents the comparison results in detail. Specifically, our CCNet outperforms recent algorithms IRNeXt and OKNet by 1.0 dB and 0.3 dB PSNR, respectively, and achieves the highest 0.99 SSIM among these methods. In addition, our CCNet achieves the 4.54 dB PSNR gain and 0.03 SSIM gain over Transformer model MSP-Former. Then, we illustrate the visual comparison results in Fig \ref{fig_9}, which demonstrates that our method can produce clearer and more natural visual results when dealing with big snowflakes, and preserve the details of snowflake-like objects, thanks to the captured abundant contextual information.

\begin{table}[htbp]
  \centering
  \caption{ Image desnowing results on CSD. CCNet outperforms other methods significantly. The best
and the second best results are hightlighted and underlined, respectively.
}
    \begin{tabular}{p{6.545em}ll}
    \toprule
    
    Method & \multicolumn{1}{p{4.045em}}{PSNR} & \multicolumn{1}{p{4.045em}}{SSIM} \\
    \midrule
    DesnowNet & 20.13 & \ 0.81 \\
    CycleGAN & 20.98 & \ 0.80 \\
    All in One & 26.31 & \ 0.87 \\
    JSTASR & 27.96 & \ 0.88 \\
    HDCW-Net & 29.06 & \ 0.91 \\
    TransWeather & 31.76 & \ 0.93 \\
    MSP-Former & 33.75 & \ 0.96 \\
    NAFNet & 35.13 & \ 0.97 \\
    SANet & 36.39 & \ \underline{0.98} \\
    IRNeXt & 37.29 & \ \textbf{0.99} \\
    OKNet & \underline{37.99} & \ \textbf{0.99} \\
    \midrule
    CCNet (ours) & \textbf{38.29} & \ \textbf{0.99} \\
    \bottomrule
    \end{tabular}%
  \label{tab3}%
\end{table}%

\subsection{Ablation studies}
A series of ablation experiments for image dehazing are conducted to analyze the efficiency of the proposed CCNet and its components, i.e., efficient residual star module (ERSM) and large dynamic integration module (LDIM). Initially, we ultilize the model without ERSM, D-RSM and LDIM as the baseline to train on the RESIDE \citep{ref22} dataset and test on the SOTS-Indoor \citep{ref22}. The tested performance 38.40 dB PSNR and 0.9944 SSIM are then recorded as the performance benchmark.

Then, under the same datasets and setting, we carry out experiments to train and test the model with the D-RSM, ERSM or LDIM. The model with the D-RSM only achieves 37.86 PSNR which is even lower than 38.40 dB PSNR of performance benchmark. Then, the model with the ERSM obtains 40.23 PSNR which is higher than 37.86 PSNR of the model with the D-RSM, demonstrating the efficiency of our ERSM and the significance of contextual information for star operation. Note that D-RSM and ERSM have the same number of parameters, and only the location of the depth-wise convolution layer is different. The 40.76 dB PSNR is obtained by the model with the LDIM which outperforms the baseline model by 2.36 dB PSNR, demonstrating the efficiency of the proposed LDIM. Notably, LDIM can substantially improve the performance of the benchmark model in comparison to ERSM, thereby indicating that LDIM is more effective than ERSM. Besides, the results demonstrate that the contextual information is crucial for image dehazing task.

Next, we train and test the model with both D-RSM and LDIM, and it achieves 39.27 dB PSNR. We train and test the model with both ERSM and LDIM, i.e., CCNet, under the same datasets and setting. The obtained 41.25 dB PSNR is the highest performance which is 2.85 dB higher than the baseline model's 38.40 dB. All experimental results are recorded in Table \ref{tab4}, and they demonstrate that ERSM and LDIM can significantly improve the model performance, and our CCNet is efficient for image dehazing due to the captured rich contextual information. 

\begin{table*}[htbp]
  \centering
  \caption{Ablation studies on SOTS-Indoor dataset. The best results are \textbf{highlighted}.}
    \begin{tabular}{p{4.045em}p{4.045em}p{4.045em}p{4.045em}cccc}
    \toprule
    Baseline & D-RSM & ERSM  & LDIM  & \multicolumn{1}{p{4.045em}}{Params/M} & \multicolumn{1}{p{4.045em}}{FLOPs/G} & \multicolumn{1}{p{4.045em}}{PSNR} & \multicolumn{1}{p{4.045em}}{SSIM} \\
    \centering\Checkmark & \centering\XSolidBrush & \centering\XSolidBrush & \centering\XSolidBrush & 4.29  & 41.91 & 38.4  & 0.9944 \\
    \centering\Checkmark & \centering\Checkmark & \centering\XSolidBrush & \centering\XSolidBrush & 4.24  & 43.47 & 37.86 & 0.9943 \\
    \centering\Checkmark & \centering\XSolidBrush & \centering\Checkmark & \centering\XSolidBrush & 4.24  & 43.47 & 40.23 & 0.9956 \\
    \centering\Checkmark & \centering\XSolidBrush & \centering\XSolidBrush & \centering\Checkmark & 4.31  & 41.94 & 40.76 & 0.9958 \\
    \centering\Checkmark & \centering\Checkmark & \centering\XSolidBrush & \centering\Checkmark & 4.26  & 43.51 & 39.27 & 0.995 \\
    \centering\Checkmark & \centering\XSolidBrush & \centering\Checkmark & \centering\Checkmark & 4.26  & 43.51 & \textbf{41.25} & \textbf{0.9963} \\
    \bottomrule
    \end{tabular}%
  \label{tab4}%
\end{table*}%

\subsection{Model Complexity Analysis}
\begin{figure}[!t]
\centering
\includegraphics[width=3in]{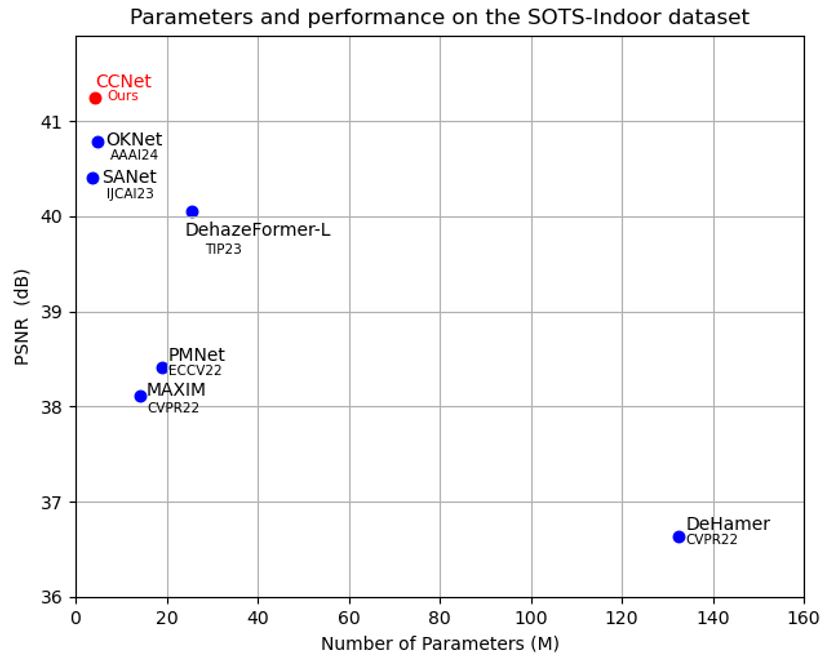}
\caption{Performance and the parameters of different methods on SOTS-Indoor dataset.}
\label{fig_10}
\end{figure}

\begin{figure}[!t]
\centering
\includegraphics[width=3in]{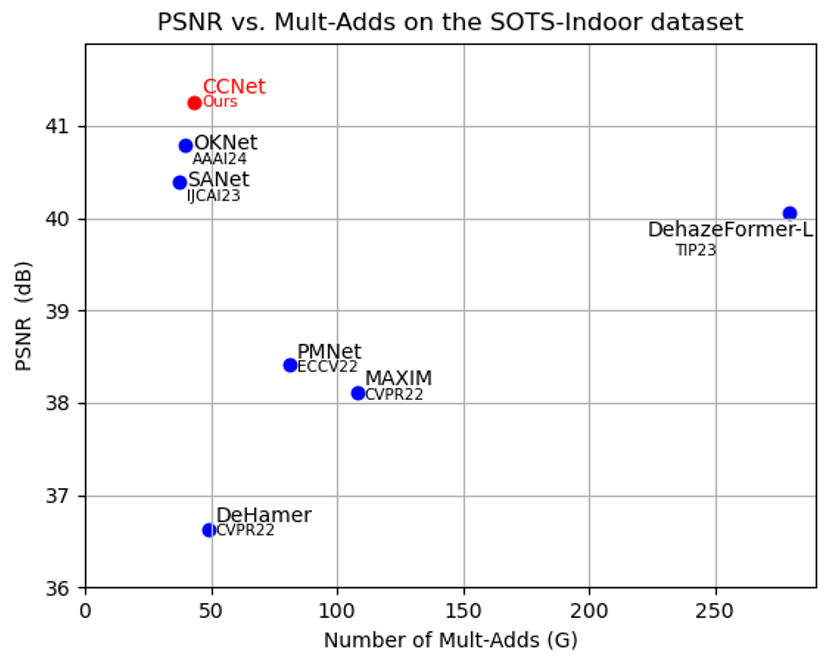}
\caption{Performance and the Multi-Adds of different methods on SOTS-Indoor dataset.}
\label{fig_11}
\end{figure}

In order to further validate the effectiveness and efficiency of our CCNet, we analyze the model complexity and make a comparison with other previous state-of-the-art algorithms, including DeHamer \citep{ref23}, MAXIM \citep{ref24}, PMNet \citep{ref25}, DehazeFormer-L \citep{ref54}, SANet \citep{ref19}, and OKNet \citep{ref21}. Specifically, on the SOTS-Indoor \citep{ref22} dataset, we conduct analysis and comparison of model parameters, model's multi-adds, and model performance. The detailed comparison results are shown in Table \ref{tab5}. It can be seen that our model with the second fewest model parameters, achieves the highest PSNR compared with recent image restoration (IR) methods. Notably, compared to recent transformer-based algorithm DehazeFormer-L, our CCNet achieves a 1.2 dB PSNR gain with one-sixth of its parameters, which demonstrates the efficiency of our proposed method. We further visually illustrate the difference among these methods in terms of parameters and PSNR in Fig \ref{fig_10}, and their difference about multi-adds and PSNR is visually illustrated in Fig \ref{fig_11}.

\begin{table*}[htbp]
  \centering
  \caption{Computation and parameters comparison for image dehazing (SOTS-Outdoor dataset). The best and the second best results are hightlighted and underlined, respectively.}
    \begin{tabular}{p{6.055em}cccccc}
    \toprule
    Metric & \multicolumn{1}{p{5.045em}}{\textcolor[rgb]{ .137,  .122,  .125}{\ DeHamer }} & \multicolumn{1}{p{4.045em}}{MAXIM } & \multicolumn{1}{p{6.945em}}{DehazeFormer-L} & \multicolumn{1}{p{4.045em}}{\ SANet} & \multicolumn{1}{p{4.045em}}{\ OKNet} & \multicolumn{1}{p{5.545em}}{CCNet (ours)} \\
    \midrule
    \textcolor[rgb]{ .137,  .122,  .125}{Paras (M)} & 132.45 & \textcolor[rgb]{ .137,  .122,  .125}{14.1} & \textcolor[rgb]{ .137,  .122,  .125}{25.44} & \textbf{3.81} & 4.72  & \underline{4.26} \\
    \textcolor[rgb]{ .137,  .122,  .125}{Mult-Adds (G)} & \textcolor[rgb]{ .137,  .122,  .125}{48.93} & \textcolor[rgb]{ .137,  .122,  .125}{108} & \textcolor[rgb]{ .137,  .122,  .125}{279.7} & \textbf{37.26} & \underline{39.67} & 43.51 \\
    PSNR (dB) & \textcolor[rgb]{ .137,  .122,  .125}{36.63} & \textcolor[rgb]{ .137,  .122,  .125}{38.11} & \textcolor[rgb]{ .137,  .122,  .125}{40.05} & 40.4  & \underline{40.79} & \textbf{41.25} \\
    \bottomrule
    \end{tabular}%
  \label{tab5}%
\end{table*}%

\section{CONCLUSION}
In this paper, we propose a context-aware convolutional network (CCNet) for efficient image restoration via the powerful learning ability for contextual high-dimensional mapping and abundant contextual information integration. Specifically, we propose an efficient residual star module (ERSM) which includes context-aware "star operation" (element-wise multiplication) to contextually map features into exceedingly high-dimensional and non-linear feature spaces. ERSM helps to capture context-aware features which effectively improve model learning ability. To further boost the extraction of contextual information, we propose a large dynamic integration module (LDIM) which possesses an extremely large square receptive field to dynamically and efficiently integrate more contextual information, and it helps to further significantly improve the reconstruction performance. Moreover, the LDIM adopts multi-scale kernels to learn vital multi-scale information to deal with image degradation of different sizes. Extensive experiments show that our CCNet outperforms state-of-the-art methods on several image restoration tasks.

\end{document}